\pdfoutput=1

\documentclass[11pt]{article}
\usepackage{coling2020}
\usepackage{times}
\usepackage{url}
\usepackage{latexsym}
\usepackage{balance}
\usepackage{bm}
\usepackage{amssymb}
\usepackage{amsmath}
\usepackage{url}
\usepackage{makecell}
\usepackage{multirow}
\usepackage{subfig}
\usepackage{graphicx} 
\usepackage{color}
\usepackage{colortbl}
\usepackage{textcomp}
\usepackage{CJK}
\usepackage{enumitem}
\usepackage{booktabs}
\usepackage{microtype}
\usepackage{arydshln}
\usepackage[ruled,linesnumbered]{algorithm2e}
\usepackage{wrapfig}

\makeatletter
\g@addto@macro\normalsize{%
  \abovedisplayskip 4pt plus 2pt minus 3pt%
  \belowdisplayskip \abovedisplayskip
  \abovedisplayshortskip 4pt plus2pt  minus3pt%
  \belowdisplayshortskip 4pt plus2pt minus3pt%
}
\usepackage{arydshln}

\colingfinalcopy 

\title{Toward Efficient Language Model Pretraining and Downstream Adaptation via Self-Evolution: A Case Study on SuperGLUE}

\author{
\small
Qihuang Zhong$^{1,2 *}$
\ Liang Ding$^{2 *}$ 
\ Yibing Zhan$^{2}$
\ Yu Qiao$^{3}$
\ Yonggang Wen$^{4}$
\ Li Shen$^{2}$
\ Juhua Liu$^{1}$
\\ 
\ \textbf{ \small
\ Baosheng Yu$^{5}$,
\ Bo Du$^{1}$, 
\ Yixin Chen$^{6}$,
\ Xinbo Gao$^{7}$,
\ Chunyan Miao$^{4}$,
\ Xiaoou Tang$^{3}$
\ Dacheng Tao$^{2}$} \\
\small
\ $^{1}$Wuhan University 
\ $^{2}$JD Explore Academy, JD.com Inc. 
\ $^{3}$Shanghai AI Lab
\ $^{4}$Nanyang Technological University
\\
\small
\ $^{5}$The University of Sydney
\ $^{6}$Washington University in St Louis
\ $^{7}$Chongqing University of Posts and Telecommunications
\\
\small
\includegraphics[scale=0.11]{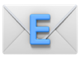} \texttt{zhongqihuang@whu.edu.cn}, \texttt{dingliang1@jd.com}
}

\date{}

\begin{document}
\maketitle


\blfootnote{
    \hspace{-0.2cm}  
    * 
    Equal contribution. Work was done when Qihuang was interning at JD Explore Academy.
}

\begin{abstract}
    This technical report briefly describes our JDExplore d-team's \textbf{Vega v2} submission on the SuperGLUE leaderboard\footnote{\url{https://super.gluebenchmark.com/leaderboard/}}. SuperGLUE is more challenging than the widely used general language understanding evaluation (GLUE) benchmark, containing eight difficult language understanding tasks, including question answering, natural language inference, word sense disambiguation, coreference resolution, and reasoning.
    \texttt{\bf [Method]}
    Instead of arbitrarily increasing the size of a pretrained language model (PLM), our aim is to 1) fully extract knowledge from the input pretraining data given a certain parameter budget, e.g., 6B, and 2) effectively transfer this knowledge to downstream tasks. To achieve goal 1), we propose self-evolution learning for PLMs to wisely predict the informative tokens that should be masked, and supervise the masked language modeling (MLM) process with rectified smooth labels. For goal 2), we leverage the prompt transfer technique to improve the low-resource tasks by transferring the knowledge from the foundation model and related downstream tasks to the target task. 
    \texttt{\bf [Results]}
    According to our submission record (Oct. 2022), with our optimized pretraining and fine-tuning strategies, our 6B Vega method achieved new state-of-the-art performance on 4/8 tasks, sitting atop the SuperGLUE leaderboard on Oct. 8, 2022, with an average score of 91.3. 
    \end{abstract}
    
    \begin{figure}[htb]
        \centering
    \includegraphics[width=0.44\textwidth]{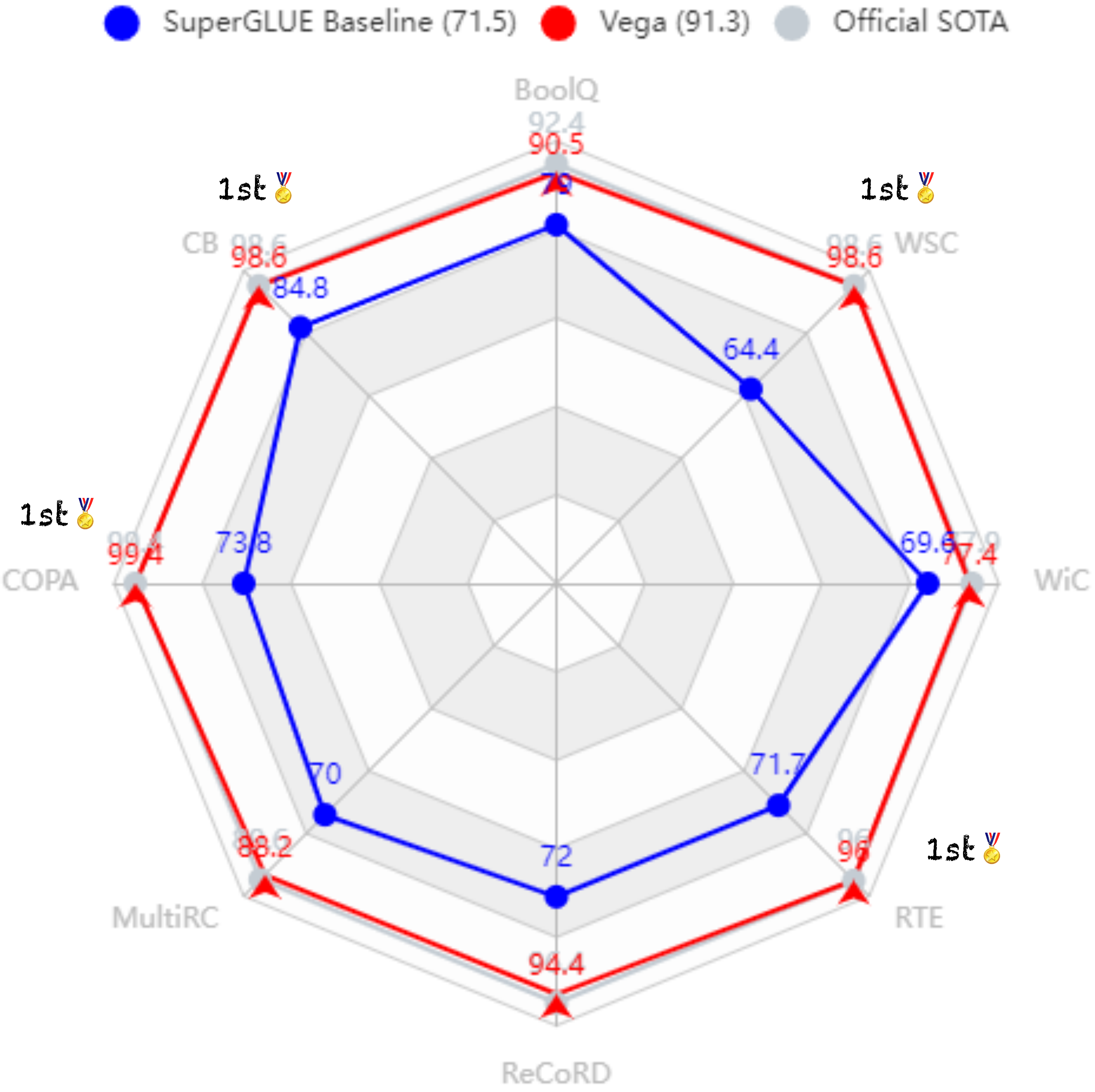}
        \caption{Vega v2 achieves state-of-the-art records on 4 out of 8 tasks among all submissions, producing the best average score of 91.3 and significantly outperforming the competitive official SuperGLUE Baseline \cite[BERT++]{superglue}.}
        \label{fig:radia}
    \end{figure}

\section{Introduction}
\label{sec:intro}

The last several years have witnessed notable progress across many natural language processing (NLP) tasks, led by pretrained language models (PLMs) such as bidirectional encoder representations from transformers (BERT)~\cite{devlin-etal-2019-bert}, OpenAI GPT \cite{radford2019language} and its most renowned evolution GPT3~\cite{GPT3}. The unifying theme of the above methods is that they conduct self-supervised learning with massive easy-to-acquire unlabelled text corpora during the pretraining stage and effectively fine-tune on a few labeled data in target tasks. 
Such a ``pretraining-fine-tuning'' paradigm has been widely adopted by academia and industry, and the main research and development direction involves scaling the sizes of foundation models up to extremely large settings, such as Google's 540B PaLM~\cite{Chowdhery2022PaLMSL}, to determine the upper capacity bounds of foundation models.

In such a context, the SuperGLUE~\cite{superglue} (a more challenging version of the general language understanding evaluation  (GLUE) benchmark~\cite{glue}) has become the most influential and prominent evaluation benchmark for the foundation model community. Most high-performing models on the GLUE/ SuperGLUE leaderboard bring new insights and better practices to properly guide future research and applications.

We recently submitted our 6B \textbf{Vega} v2
model to the SuperGLUE leaderboard and, as seen in Figure~\ref{fig:radia}, obtained state-of-the-art records on 4 out of 8 tasks, sitting atop the leaderboard as of Oct. 8, 2022, with an average score of 91.3. 
Encouragingly, our 6B model with deliberately optimized pretraining and downstream adaptation strategies substantially outperforms 540B PaLM~\cite{Chowdhery2022PaLMSL}, showing the effectiveness and parameter-efficiency of our Vega model.
This technical report briefly describes how we build our powerful model under a certain parameter budget, i.e., 6B, from different aspects, including backbone framework (\S\ref{subsec:backbone}), the efficient pretraining process (\S\ref{subsec:pretrain}), and the downstream adaptation approach (\S\ref{subsec:downstream}). To fully extract knowledge from the given pretraining data to PLMs, we propose a \textbf{self-evolution learning} (in Figure~\ref{fig:self-evolution}) mechanism to wisely predict the informative tokens that should be masked and supervise the mask language modeling process with rectified smooth labels. To effectively transfer the knowledge to different downstream tasks, especially the low-resource tasks, e.g., CB, COPA, and WSC, we design a \textbf{knowledge distillation-based prompt transfer} method~\cite{zhong2022panda} (in Figure~\ref{fig_panda}) to achieve better performance with improved robustness.

The remainder of this paper is designed as follows. We introduce the major utilized approaches in Section~\ref{sec:app}. In Section~\ref{sec:exp}, we review the task descriptions and data statistics and present the experimental settings and major results. Conclusions are described in Section~\ref{sec:con}.

\section{Approaches}
\label{sec:app}
In this section, we describe the main techniques in our model, including the backbone framework in~\S\ref{subsec:backbone}, the efficient pretraining approaches in~\S\ref{subsec:pretrain}, and the downstream adaptation technique in~\S\ref{subsec:downstream}.

\subsection{Backbone Framework}
\label{subsec:backbone}
Vanilla transformers~\cite{transformer} enjoy appealing scalability as large-scale PLM backbones~\cite{devlin-etal-2019-bert,t5,GPT3,zan2022vegamt}; for example, T5 and GPT3 flexibly scale their feedforward dimensions and layers up to 65,534 and 96, respectively.
We hereby employ a vanilla transformer, i.e., multihead self-attention followed by a fully connected feedforward network, as our major backbone framework. 
As encoder-only PLMs have an overwhelming advantage over the existing methods on the SuperGLUE leaderboard, we train our large model in an encoder-only fashion to facilitate downstream language understanding tasks. 
According to our PLM parameter budget -- 6 Billion, we empirically set the model as follows: 24 layers, 4096 as the hidden layer size, an FFN of size 16,384, 32 heads, and 128 as the head size.
In addition, \newcite{deberta} empirically demonstrated the necessity of computing self-attention with disentangled matrices based on their contents and relative positions, namely disentangled attention\footnote{In our preliminary ablations, we surprisingly found the Enhanced Mask Decoder~\cite{deberta} technique, which is coupled with disentangled attention strategy in DeBRETa, was useless, therefore we did not this strategy in Vega v2.}, which is adopted in Vega v2.

\subsection{Efficient Pretraining}
\label{subsec:pretrain}
Recall that our aim is not to arbitrarily increase the model scales, but to facilitate storing the informative derived knowledge from the pretraining data in PLMs.
To approach this goal, we first revisit the representative self-supervision objective -- masked language modeling~\cite{devlin-etal-2019-bert} (MLM), and propose a novel self-evolution learning mechanism to enable our PLM to wisely predict the informative tokens that should be masked, and train the model with smooth self-evolution labels.

\paragraph{Masked Language Modeling} MLM is a widely used self-supervision objective when conducting large-scale pretraining on large amounts of text to learn contextual word representations. In practice, MLM randomly selects a subset of tokens from a sentence and replaces them with a special mask token, i.e., \texttt{[MASK]}. However, such a random masking procedure is usually suboptimal, as the masked tokens are sometimes too easy to guess with only local cues or shallow patterns. Hence, some prior works focused on more informative masking strategies, such as span-level masking~\cite{joshi2020spanbert}, entity-level masking~\cite{sun2019ernie}, and pointwise mutual information (PMI)-based masking~\cite{sadeq2022informask}. These efforts achieved better performance than vanilla random masking, which inspires us to explore more approaches for fully extracting knowledge from pretraining data. 

\begin{figure*}[t!]
    \centering
\includegraphics[width=0.99\textwidth]{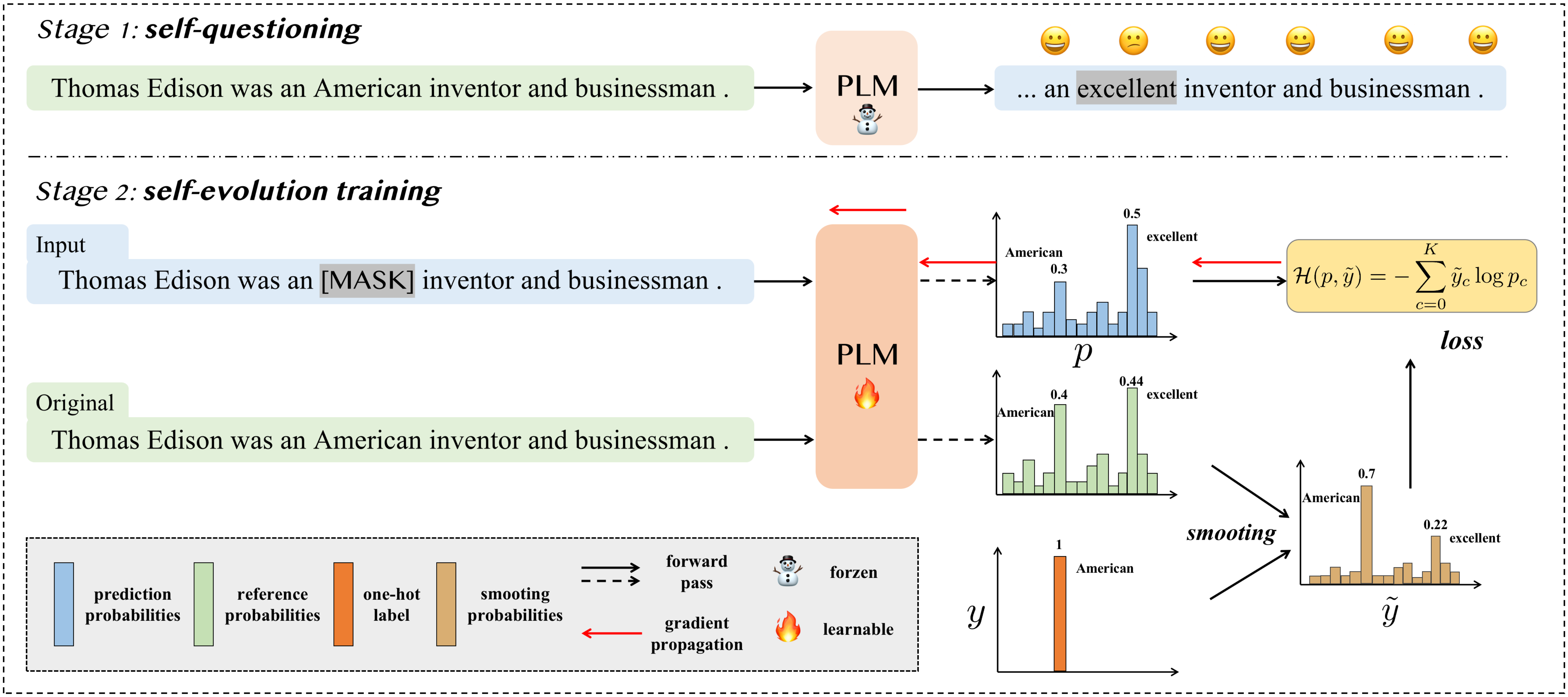}
    \caption{\label{fig:self-evolution}Overview of the proposed \textbf{self-evolution learning mechanism} for PLMs.}
\end{figure*}

\paragraph{Self-Evolution Learning}
Based on the above motivations, we propose a novel self-evolution learning mechanism for PLMs, as illustrated in Figure~\ref{fig:self-evolution}. Different from the prior works that designed masking strategies to train language models from scratch, our self-evolution learning approach aims to encourage the given ``naive'' PLMs to find patterns (tokens) that are not learned well but are more informative, and then fix them. Specifically, there are two stages in our self-evolution learning mechanism, as follows.

Stage 1 is \textit{\textbf{self-questioning}}. Given a vanilla PLM (e.g., trained with the random masking objective), we first feed the original training samples into the PLM and make it re-predict the output probabilities for each token. As the PLM has seen these samples and learned from them in the pretraining stage, it can make deterministic and correct predictions for most of the tokens, which we denote as learned tokens. However, for some tokens, such as ``\texttt{American}'' in the sentence ``\texttt{Thomas Edison was an American inventor and businessman}'', the PLM tends to predict this token as ``\texttt{excellent}'' (the probability of ``excellent'' is 0.44, while the probability of ``American'' is 0.4). We attribute this phenomenon to the fact that the PLM does not learn this knowledge-intense pattern but only makes its prediction based on the local cues. We refer to these harder and more informative tokens as neglected tokens. After all training samples are fed into the PLM, we can obtain a set of neglected tokens for each training sample. Note that this procedure is conducted offline and does not update the parameters of the original PLM.

Stage 2 is \textit{\textbf{self-evolution training}}. Given the neglected tokens (obtained in stage 1), we can select them for masking and then encourage the PLM to learn from these informative patterns, thus continuously improving the capability of the PLM. Intuitively, we can make the PLM learn how to predict these tokens, by minimizing the loss between the predicted probabilities and one-hot labels. However, considering the diversity of the neglected token, if we force the PLM to promote one specified ground truth over others, the other reasonable ``ground truths'' (for a given masking token, there can be more than one reasonable prediction) become false negatives that may plague the training process or cause a performance decrease~\cite{li2022pre}.

Hence, we propose a novel self-evolution training method to help the PLM learn from the informative tokens, without damaging the diversification ability of the PLM. In practice, we feed the masked sentence and original sentence into the PLM, and obtain the prediction probabilities $p$ and reference probabilities $r$ for the \texttt{[MASK]} token. Then, we merge the $r$ and the one-hat label $y$ as $\Tilde{y}=(1-\alpha)*y+\alpha*r$, where $\Tilde{y}$ denotes the smoothing label probabilities and $\alpha$ is a weighting factor that is empirically set as 0.5. Finally, we use the cross-entropy loss function to minimize the difference between $p$ and $\Tilde{y}$. In this way, different from the strong-headed supervision of the one-hot labels $y$, the PLM can benefit more from the smooth and informative labels $\Tilde{y}$.

\subsection{Downstream Adaptation}
\label{subsec:downstream}
In addition to the above efficient pretraining methods, we also introduce some useful fine-tuning strategies for effectively adapting our Vega v2 to downstream tasks. Specifically, there are two main problems that hinder the adaptation performance of a model. 1) The first concerns the domain gaps between the training and test sets, which lead to poor performance on target test sets. 2) The second is the use of limited training data, e.g., the CB task, which only consists of 250 training samples, as limited data can hardly update the total parameters of PLMs effectively.
Note that in addition to the strategies listed below, we have also designed and implemented other methods from different perspectives to improve the generalization and efficiency of models, e.g. the FSAM optimizer for PLMs~\cite{Zhong2022ImprovingSM}, SparseAdapter~\cite{He2022SparseAdapterAE}, and continued training with downstream data~\cite{Zan2022BridgingCG}. Although these approaches can help to some extent, they do not provide complementary benefits compared to the listed approaches, so they are not described here.

\paragraph{Transductive Fine-tuning}
\begin{algorithm}[t]
\KwIn{Finetuned (FT) Model $M$, ~~~~~~~~~~~~~~~~Downstream Seed $D$}
\KwOut{Transductively FT Model $M^{'}$}
        $t:=0$\\
        \While {not convergence}{
      Estimate $D$ with $M$ and get $D^{M}$\\
      Tune $M$ on $D \cup D^{M}$ and get $M^{'}$, then $M=M^{'}$ \\ 
        $t := t + 1$\\      
        }
\caption{Transductive Fine-tuning} 
\label{alg:1}
\end{algorithm}
Regrading the domain or linguistic style gap between the training and test sets (the first problem), we adopt a transductive fine-tuning strategy to improve the target domain performance, which is a common practice in machine translation evaluations~\cite{wu-etal-2020-tencent,ding2021usyd} and some domain adaptation applications~\cite{liu2020semitext}. 
The proposed transductive fine-tuning technique is shown in Algorithm~\ref{alg:1}. Whether we should conduct transductive fine-tuning depends on the practical downstream performance achieved.

\paragraph{Prompt-Tuning} To address the second problem, we replace the vanilla fine-tuning process with a more parameter-efficient method, \textit{prompt-tuning}~\cite{lester2021power}, for low-resource tasks. Despite the success of prompts in many NLU tasks~\cite{wang2022c3da,zhong2022e2s2}, directly using prompt-tuning might lead to poor results, as this approach is sensitive to the prompt's parameter initialization settings~\cite{zhong2022panda}. An intuitive approach, termed as prompt transfer~\cite{vu2022spot}, is to initialize the prompt on the target task with the trained prompts from similar source tasks. Unfortunately, such a vanilla prompt transfer approach usually achieves suboptimal performance, as (i) the prompt transfer process is highly dependent on the similarity of the source-target pair and (ii) directly tuning a prompt initialized with the source prompt on the target task might lead to forgetting the useful general knowledge learned from the source task.

To this end, we introduce a novel prompt transfer framework~\cite{zhong2022panda} to tackle the above problems. For (i), we propose a new metric to accurately predict prompt transferability. In practice, the metric first maps the source/target tasks into a shared semantic space to obtain their task embeddings based on the source/target soft prompts and then measures the prompt transferability via the similarity of corresponding task embeddings. In our primary experiments, we found that this metric could make appropriately choose which source tasks should be used for a target task. For instance, to perform prompt transfer among the SuperGLUE tasks, WSC is a better source task for the CB task, while COPA benefits more from the RTE task.

\begin{figure}[t]
    \centering
\includegraphics[width=0.7\textwidth]{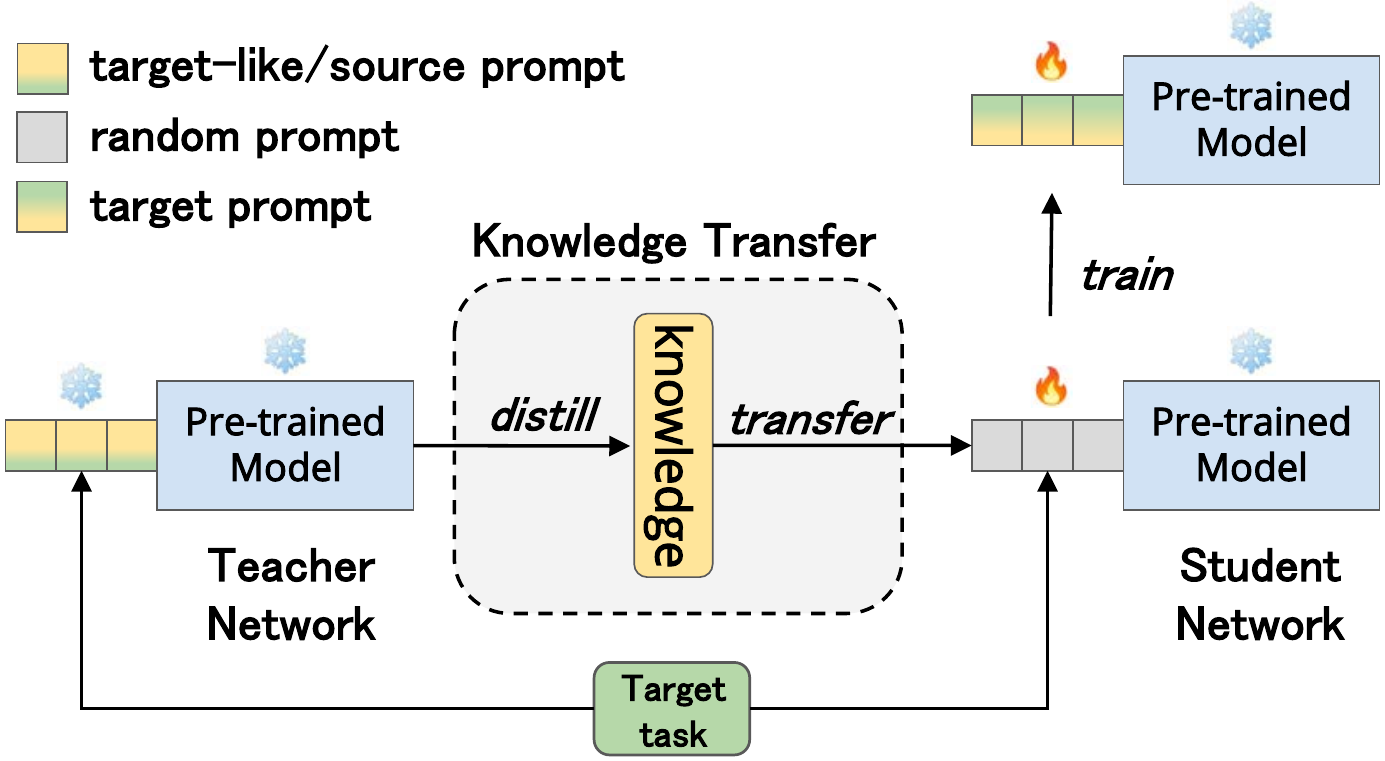}
    \caption{The architecture of our proposed KD-based prompt transfer method.}
    \label{fig_panda}
\end{figure}

Regarding (ii), inspired by the knowledge distillation (KD) paradigm~\cite{hinton2015distilling,liu2021unified,Rao2022ParameterEfficientAS} that leverages a powerful teacher model to guide the training process of a student model, we propose a KD-based prompt transfer method that leverages the KD technique to transfer the knowledge from the source prompt to the target prompt in a subtle manner, thus effectively alleviating the problem of prior knowledge forgetting. An illustration of our proposed method is shown in Figure~\ref{fig_panda}.
More specifically, our KD-based prompt transfer approach first uses the PLM with the source prompt as the teacher network and the PLM with the randomly initialized prompt as the student network. Then, the student network is trained using the supervision signals from both the ground-truth labels in the target task and the soft targets predicted by the teacher network. Note that we only update the parameters of the student prompt, while keeping the other parameters fixed. Furthermore, to adaptively control the knowledge transfer process in our approach, we use the prompt similarity predicted by our metric as the balancing factor between the two supervision signals for each source-target pair.

\paragraph{Adversarial Fine-Tuning}
In addition to the above transductive FT and prompt-tuning processed for deliberately solving the training-testing domain gap and low downstream resource problem, respectively, we also adopt the advanced adversarial fine-tuning algorithm~\cite{Miyato2019VirtualAT,Jiang2020SMARTRA} designed for PLMs, i.e., SiFT~\cite{deberta}, to improve the training stability of our approach. In practice, we follow \newcite{deberta} by applying the perturbations to the normalized word embeddings when tuning our Vega foundation model on downstream tasks, where we first normalize the embedding vectors into stochastic vectors and then apply the perturbations to the normalized embedding vectors.

\section{Experiments}
\label{sec:exp}

\subsection{Implementation}
For pretraining, we follow many prior works~\cite{liu2019roberta,deberta} and use Wikipedia\footnote{\url{https://dumps.wikimedia.org/enwiki/}} (the English Wikipedia dump, 10 GB), BookCorpus~\cite{zhu2015aligning}\footnote{\url{https://github.com/butsugiri/homemade_bookcorpus}} (6 GB), OpenwebText\footnote{\url{http://Skylion007.github.io}} (38 GB), Stories\footnote{\url{https://github.com/tensorﬂow/models/tree/master/research/lm_commonsense}} (31 GB) and CC-News~\cite{trinh2018simple} (76 GB) as pretraining datasets, and use 40 NVIDIA DGX nodes with 320 A100 GPUs to train our Vega v2 model. It takes 30 days to finish phase-1 (pretraining, i.e., MLM) with 1 M steps. For phase-2, i.e., self-evolution training, we continuously train Vega v2 for 50K steps. 
During fine-tuning, we only apply our KD-based prompt transfer strategy to the low-resource tasks, e.g., CB, COPA, and WSC. For the other tasks, the vanilla full-parameter model-tuning method with adversarial and transductive fine-tuning is used. We use AdamW~\cite{loshchilov2018decoupled} as the optimizer for both pretraining and fine-tuning.

\subsection{Tasks}
To validate the effectiveness of Vega v2, we use the SuperGLUE benchmark~\cite{wang2019superglue} for model evaluation purposes. As one of the most popular NLU benchmarks, SuperGLUE consists of eight challenging NLU tasks, including question answering (BoolQ,~\newcite{clark2019boolq}, MultiRC,~\newcite{khashabi2018looking}, ReCoRD,~\newcite{zhang2018record}), natural language inference (CB,~\newcite{demarneffe:cb}, RTE,~\newcite{dagan2006pascal,bar2006second,giampiccolo2007third,bentivogli2009fifth}), word sense disambiguation (WIC,~\newcite{pilehvar2018wic}), coreference resolution (WSC,~\newcite{levesque2012winograd}), and reasoning (COPA,~\newcite{roemmele2011choice}). More detailed data statistics and examples for the above tasks can be found in Appendix Tables~\ref{tab:tasks} and ~\ref{tab:examples}.

\begin{table*}[!t]
\caption{\textbf{Results obtained on the SuperGLUE test sets}, which are scored by the SuperGLUE evaluation server. We obtained the results from \url{https://super.gluebenchmark.com} on October 8, 2022. The best results (except the those of human baselines) are shown in bold.}
\label{table_main}
\centering
\scalebox{0.76}{
\begin{tabular}{lcccccccccccc}
\toprule
\multirow{2}{*}{\textbf{Model}}    & BoolQ & \multicolumn{2}{c}{CB} & COPA  & \multicolumn{2}{c}{MultiRC} & \multicolumn{2}{c}{ReCoRD} & RTE  & WiC  & WSC   & \multirow{2}{*}{Score} \\ \cmidrule{2-12}
                          & Acc   & F1         & Acc       & Acc   & F1           & EM           & F1           & Acc.        & Acc  & Acc  & Acc   &                        \\ \midrule \midrule
\textbf{SuperGLUE Baselines}       & 79.0  & 84.8       & 90.4      & 73.8  & 70.0         & 24.1         & 72.0         & 71.3        & 79.0 & 69.6 & 64.4  & 71.5                   \\
\textbf{SuperGLUE Human Baselines} & 89.0  & 95.8       & 98.9      & 100.0 & 81.8         & 51.9         & 91.7         & 91.3        & 93.6 & 80.0 & 100.0 & 89.8                   \\ \hdashline
\textbf{PaLM 540B}~\cite{chowdhery2022palm}                 & 91.9  & 94.4       & 96.0      & 99.0  & 88.7         & 63.6         & 94.2         & 93.3        & 94.1 & 77.4 & 95.9  & 90.4                   \\
\textbf{ERNIE 3.0}~\cite{sun2021ernie}                 & 91.0  & 98.6       & \textbf{99.2}      & 97.4  & 88.6         & 63.2         & 94.7         & 94.2        & 92.6 & 77.4 & 97.3  & 90.6                   \\
\textbf{Turing NLR v5}~\cite{bajaj2022metro}             & 92.0  & 95.9       & 97.6      & 98.2  & 88.4         & 63.0         & \textbf{96.4}         & \textbf{95.9}        & 94.1 & 77.1 & 97.3  & 90.9                   \\
\textbf{ST-MoE-32B}~\cite{zoph2202st}                & \textbf{92.4}  & 96.9       & 98.0      & 99.2  & \textbf{89.6}         & \textbf{65.8}         & 95.1         & 94.4        & 93.5 & \textbf{77.7} & 96.6  & 91.2                   \\
\textbf{Vega v2 (Ours)}               & 90.5  & \textbf{98.6}       & \textbf{99.2}      & \textbf{99.4}  & 88.2         & 62.4         & 94.4         & 93.9        & \textbf{96.0} & 77.4 & \textbf{98.6}  & \textbf{91.3}     \\
\bottomrule
\end{tabular}
}
\end{table*}

\subsection{Main Results}
Table~\ref{table_main} reports the test results obtained by our Vega v2 and other cutting-edge models on the SuperGLUE benchmark\footnote{We show the detailed ranking results on the SuperGLUE leaderboard in the Appendix. Please refer to Table~\ref{tab:allrank}.}. Vega v2 significantly surpasses the powerful human baselines in terms of average score (91.3 \textit{vs.} 89.8) and achieves state-of-the-art performance on four (relatively) low-resource tasks, i.e., CB, COPA, RTE, and WSC. We attribute this success to the novel self-evolution learning mechanism and KD-based prompt transfer method. More specifically, the former enhances Vega v2's ability to extract informative patterns, while the latter alleviates the problem of overfitting and boosts the model performance on low-resource tasks.

In addition, compared to the other larger PLMs, e.g., PaLM~\cite{Chowdhery2022PaLMSL} which consists of 540 billion parameters, our 6-billion-parameter Vega v2 can achieve competitive or even better performance on the SuperGLUE benchmark. This inspires us to conclude that scaling PLMs to larger model sizes arbitrarily might not be cost-effective, but would encourage the PLMs to fully extract knowledge from the pretraining data when given a certain parameter budget.

\section{Conclusion}
\label{sec:con}
This paper presents the JD Explore Academy large-scale Vega v2 PLM for the SuperGLUE benchmark. 
Based on an advanced transformer backbone with disentangled attention and a series of advanced fine-tuning strategies, we propose two novel techniques. 
The first is a self-evolution learning mechanism that fully exploits the knowledge contained in data for a PLM in two steps: 1) the PLM performs self-questioning to determine hard and informative words, and then 2) supervises the MLM process with rectified smooth labels.
The second is a prompt transfer strategy for efficiently adapting downstream tasks (especially low-resource tasks) by leveraging the knowledge acquired from the foundation model and related downstream tasks.

We show that these techniques significantly improve the efficiency of model pretraining and the performance achieved on downstream tasks. The Vega v2 model with 6 billion parameters achieves state-of-the-art records on 4 out of 8 tasks and ranks the first in terms of the macro-average score.
Our experience with building Vega v2 demonstrates the necessity of 1) fully improving the parameter efficiency of PLMs, and 2) wisely preforming downstream adaptation.

\section*{Acknowledgments}
The authors wish to thank the leaderboard maintainer of SuperGLUE for their great construction efforts and their prompt responses to our questions. The authors also especially thank Mr. Yukang Zhang (JD Explore Academy), who kindly supports maintaining a stable computing platform.

\bibliographystyle{coling}
\bibliography{coling2020}

\appendix
\begin{table*}[h]
\caption{\textbf{Results of Top-10 models on SuperGLUE leaderboard} (\url{https://super.gluebenchmark.com/leaderboard}), on October 8, 2022.
}
\label{tab:allrank}
\centering \small
\scalebox{0.85}{
\begin{tabular}{clcccccccccccc}
\toprule
\multirow{2}{*}{\bf $\#$Rank} & \multicolumn{1}{c}{\multirow{2}{*}{\bf Model}} & \textbf{BoolQ} & \multicolumn{2}{c}{\bf CB} & \textbf{COPA}  & \multicolumn{2}{c}{\bf MultiRC} & \multicolumn{2}{c}{\bf ReCoRD} & \textbf{RTE}  & \textbf{WiC}  & \textbf{WSC}   & \multirow{2}{*}{\bf Score} \\ \cmidrule{3-13}
                      & \multicolumn{1}{c}{}                       & Acc   & F1         & Acc       & Acc   & F1           & EM           & F1           & Acc.        & Acc  & Acc  & Acc   &                        \\ \midrule
1                     & Vega v2 (Ours)                                & 90.5  & 98.6       & 99.2      & 99.4  & 88.2         & 62.4         & 94.4         & 93.9        & 96.0 & 77.4 & 98.6  & 91.3                   \\
2                     & ST-MoE-32B                                 & 92.4  & 96.9       & 98.0      & 99.2  & 89.6         & 65.8         & 95.1         & 94.4        & 93.5 & 77.7 & 96.6  & 91.2                   \\
3                     & Turing NLR v5                              & 92.0  & 95.9       & 97.6      & 98.2  & 88.4         & 63.0         & 96.4         & 95.9        & 94.1 & 77.1 & 97.3  & 90.9                   \\
4                     & ERNIE 3.0                                  & 91.0  & 98.6       & 99.2      & 97.4  & 88.6         & 63.2         & 94.7         & 94.2        & 92.6 & 77.4 & 97.3  & 90.6                   \\
5                     & PaLM 540B                                  & 91.9  & 94.4       & 96.0      & 99.0  & 88.7         & 63.6         & 94.2         & 93.3        & 94.1 & 77.4 & 95.9  & 90.4                   \\
6                     & T5+UKG, Single Model        & 91.4  & 95.8       & 97.6      & 98.0  & 88.3         & 63.0         & 94.2         & 93.5        & 93.0 & 77.9 & 96.6  & 90.4                   \\
7                     & DeBERTa/ TuringNLRv4                       & 90.3  & 95.7       & 97.6      & 98.4  & 88.2         & 63.7         & 94.5         & 94.1        & 93.2 & 77.5 & 95.9  & 90.3                   \\
8                     & SuperGLUE Human Baselines                  & 89.0  & 95.8       & 98.9      & 100.0 & 81.8         & 51.9         & 91.7         & 91.3        & 93.6 & 80.0 & 100.0 & 89.8                   \\
9                     & T5                                         & 91.2  & 93.9       & 96.8      & 94.8  & 88.1         & 63.3         & 94.1         & 93.4        & 92.5 & 76.9 & 93.8  & 89.3                   \\
10                    & Frozen T5 1.1 + SPoT                       & 91.1  & 95.8       & 97.6      & 95.6  & 87.9         & 61.9         & 93.3         & 92.4        & 92.9 & 75.8 & 93.8  & 89.2                \\
\bottomrule
\end{tabular}
}
\end{table*}

\begin{table*}[t]
\caption{\textbf{Data statistics of different tasks included in SuperGLUE} according to their original paper~\cite{superglue}.  
\textit{WSD} stands for word sense disambiguation, \textit{NLI} is natural language inference, \textit{coref.} is coreference resolution, and \textit{QA} is question answering. For MultiRC, we list the number of total answers for 456/83/166 train/dev/test questions. 
}
\centering \small
\begin{tabular}{lrrrlll}
 \toprule
\textbf{Corpus} & \textbf{$|$Train$|$} & \textbf{$|$Dev$|$} & \textbf{$|$Test$|$} & \textbf{Task} & \textbf{Metrics} & \textbf{Text Sources} \\
\midrule 
BoolQ & 9427 & 3270 & 3245 & QA & acc. & Google queries, Wikipedia \\
CB & 250 & 57 & 250 & NLI & acc./F1 & various \\
COPA & 400 & 100 & 500 & QA & acc. & blogs, photography encyclopedia\\
MultiRC & 5100 & 953 & 1800 & QA & F1$_a$/EM & various \\
ReCoRD & 101k & 10k & 10k & QA & F1/EM & news (CNN, Daily Mail) \\
RTE & 2500 & 278 & 300 & NLI & acc. & news, Wikipedia \\
WiC & 6000 & 638 & 1400 & WSD & acc. & WordNet, VerbNet, Wiktionary \\
WSC & 554 & 104 & 146 & coref. & acc. & fiction books \\
\bottomrule
\end{tabular}
\label{tab:tasks}
\end{table*}

\begin{table*}[t]
\caption{\textbf{Task examples from the valid set in SuperGLUE}~\cite{superglue}. \textbf{Bold} text represents part of the example format for each task. Text in \textit{italics} is part of the model input. \underline{\textit{Underlined}} text is specially marked in the input. Text in a \texttt{monospaced font} represents the expected model output.}
\label{tab:examples}
\centering \footnotesize
\begin{tabular}{p{0.005\textwidth}p{0.93\textwidth}}
 \toprule
 \parbox[t]{1mm}{\multirow{2}{*}{\rotatebox[origin=c]{90}{{\textbf{BoolQ}}}}} &
\textbf{Passage:} \textit{Barq's -- Barq's is an American soft drink. Its brand of root beer is notable for having caffeine. Barq's, created by Edward Barq and bottled since the turn of the 20th century, is owned by the Barq family but bottled by the Coca-Cola Company. It was known as Barq's Famous Olde Tyme Root Beer until 2012.} \\ & \textbf{Question:} \textit{is barq's root beer a pepsi product} \quad \textbf{Answer:} \texttt{No}\\

\midrule
\parbox[t]{1mm}{\multirow{2}{*}{\rotatebox[origin=c]{90}{{\textbf{CB}}}}} &
\textbf{Text:} \textit{B: And yet, uh, I we-, I hope to see employer based, you know, helping out. You know, child, uh, care centers at the place of employment and things like that, that will help out. A: Uh-huh. B: What do you think, do you think we are, setting a trend?} \\ & \textbf{Hypothesis:} \textit{they are setting a trend} \quad \textbf{Entailment:} \texttt{Unknown}\\

\midrule
\parbox[t]{1mm}{\multirow{2}{*}{\rotatebox[origin=c]{90}{{\textbf{COPA}}}}} & \textbf{Premise:} \textit{My body cast a shadow over the grass.}\quad
\textbf{Question:} \textit{What’s the CAUSE for this?}\\
&\textbf{Alternative 1:} \textit{The sun was rising.}\quad
\textbf{Alternative 2:} \textit{The grass was cut.}\\
&\textbf{Correct Alternative:} \texttt{1}\\

\midrule
\parbox[t]{1mm}{\multirow{2}{*}{\rotatebox[origin=c]{90}{{\textbf{MultiRC}}}}} &
\textbf{Paragraph:} \textit{Susan wanted to have a birthday party. She called all of her friends. She has five friends. Her mom said that Susan can invite them all to the party. Her first friend could not go to the party because she was sick. Her second friend was going out of town. Her third friend was not so sure if her parents would let her. The fourth friend said maybe. The fifth friend could go to the party for sure. Susan was a little sad. On the day of the party, all five friends showed up. Each friend had a present for Susan. Susan was happy and sent each friend a thank you card the next week}\\
& \textbf{Question:} \textit{Did Susan's sick friend recover?} \textbf{Candidate answers:} 
\textit{Yes, she recovered} (\texttt{T}), 
\textit{No} (\texttt{F}), 
\textit{Yes} (\texttt{T}), 
\textit{No, she didn't recover} (\texttt{F}), 
\textit{Yes, she was at Susan's party} (\texttt{T})\\

\midrule
\parbox[t]{1mm}{\multirow{2}{*}{\rotatebox[origin=c]{90}{{\textbf{ReCoRD}}}}} & 
\textbf{Paragraph:} \textit{(\underline{CNN}) \underline{Puerto Rico} on Sunday overwhelmingly voted for statehood. But Congress, the only body that can approve new states, will ultimately decide whether the status of the \underline{US} commonwealth changes. Ninety-seven percent of the votes in the nonbinding referendum favored statehood, an increase over the results of a 2012 referendum, official results from the \underline{State Electorcal Commission} show. It was the fifth such vote on statehood. "Today, we the people of \underline{Puerto Rico} are sending a strong and clear message to the \underline{US Congress}  ... and to the world ... claiming our equal rights as \underline{American} citizens, \underline{Puerto Rico} Gov. \underline{Ricardo Rossello} said in a news release. @highlight \underline{Puerto Rico} voted Sunday in favor of \underline{US} statehood}\\
&\textbf{Query} For one, they can truthfully say, ``Don't blame me, I didn't vote for them, '' when discussing the <placeholder> presidency \quad \textbf{Correct Entities:} \texttt{US} \\

\midrule
\parbox[t]{1mm}{\multirow{2}{*}{\rotatebox[origin=c]{90}{{\textbf{RTE}}}}} &
\textbf{Text:} \textit{Dana Reeve, the widow of the actor Christopher Reeve, has died of lung cancer at age 44, according to the Christopher Reeve Foundation.}\\
& \textbf{Hypothesis:} \textit{Christopher Reeve had an accident.} \quad
\textbf{Entailment:} \texttt{False}\\

\midrule
\parbox[t]{1mm}{\multirow{2}{*}{\rotatebox[origin=c]{90}{{\textbf{WiC}}}}} &
\textbf{Context 1:} \textit{Room and \underline{board}.} \quad
\textbf{Context 2:} \textit{He nailed \underline{boards} across the windows.} \\
& \textbf{Sense match:} \texttt{False}\\

\midrule
\parbox[t]{1mm}{\multirow{1}{*}{\rotatebox[origin=c]{90}{{\textbf{WSC}}}}} & 
\textbf{Text:} \textit{Mark told \underline{Pete} many lies about himself, which Pete included in his book. \underline{He} should have been more truthful.} \quad \textbf{Coreference:} \texttt{False}\vspace{0.25em}\\
\bottomrule
\end{tabular}

\end{table*}

\end{document}